\title{Automated Adversarial Collaboration for \\ Advancing Theory Building in the Cognitive Sciences}
\documentclass[preprint]{ccn}


\usepackage{wrapfig}  
\usepackage{graphicx}
\usepackage{dblfloatfix}

\author{%
  Suyog Chandramouli \affmark{1} \And
  George Kachergis   \affmark{2} \AND
  Akshay K. Jagadish \affmark{3}
}

\affiliation{1}{Department of Psychology, Princeton University}
\affiliation{2}{Department of Psychology, Stanford University}
\affiliation{3}{Princeton AI Lab, Princeton University}

\emails{suyoghc@princeton.edu, george.kachergis@gmail.com, akshay.jagadish@princeton.edu}

\begin{document}
\maketitle

\maketitle

\begin{abstract}
Cognitive science often evaluates theories through narrow paradigms and local model comparisons, limiting the integration of evidence across tasks and realizations. We introduce an automated adversarial collaboration framework for adjudicating among competing theories even when the candidate models and experiments must be discovered during the adjudication process. The system combines LLM-based theory agents, program synthesis, and information-theoretic experimental design in a closed loop. In a simulation study spanning three classic categorization theories, the framework recovered the ground-truth theory across noise settings with weaker reliability in the hardest settings. Together, the framework and findings provide a concrete proof of concept for closed-loop, in-silico theory adjudication in cognitive science.
\end{abstract}


\section{Introduction}
Building better theories involves integrating evidence across broad experimental design spaces and a wide range of model classes that are aligned with specific theoretical claims; this is inevitable as theories seek to generalize across tasks and admit multiple computational realizations \citep{marr1982representation, van2021theory}. In practice, these spaces are too large to sample systematically with traditional scientific workflows \citep{musslick2025automating}. As a result, much of cognitive science remains focused on narrow paradigms and local model structures rather than the development and integration of broader theories \citep{griffiths2015manifesto}. 

Previous efforts to scale up cognitive science have concentrated on individual components of the theory-building process, such as behavioral data collection \citep{hartshorne2019thousand,jones2017bigdata}, computational modeling \citep{peterson2021using}, and optimal experiment design \citep{cavagnaro2010adaptive}. However, recent advances in automated scientific discovery methods have raised the possibility of these components coming together as a larger in-silico discovery loop \citep{musslick2025automating}. In line with this, \citet{jagadish2026can} have outlined a vision for an in-silico science of the mind in which large language model (LLM)-based systems design experiments, generate synthetic behavioral data, synthesize models, and perform critic-guided iterative refinement, bringing automated methods closer to theory development at scale. Yet, this vision leaves open the problem of theory adjudication: how to link competing theoretical accounts to executable model candidates and iteratively discriminate among them through appropriate experiments.

In this work, we take a first step towards automated theory adjudication by bringing adversarial collaboration \citep{latham1988resolving, gilovich1998varieties, mellers2001frequency} into an in-silico setting. In our framework, LLM agents represent competing theoretical accounts, instantiate them as candidate models, and propose experiments that most sharply distinguish their preferred account from rival theories. After observing the results from their experiments, the system updates its beliefs about candidate theories, each agent revises its model specification, and the process continues iteratively. This operationalizes a closed-loop procedure for adjudicating among theories when neither the candidate models nor the most informative experiments are fixed in advance. We consider three classic theories from the categorization literature \citep{ashby_human_2005}, and demonstrate how our framework can recover ground-truth theories, without requiring manual a priori specification of candidate models and experiments.

\section{Setup}

Each run (see Figure 1) begins by registering LLM-based theory agents, each associated with a theoretical claim and one or more candidate models that instantiate it.
Candidate models are currently synthesized using GeCCo \citep{rmusgenerating}, but other program synthesis methods or hand-crafted models can also be provided.

\begin{figure*}[!t]
    \centering
    \includegraphics[width=0.9\textwidth]{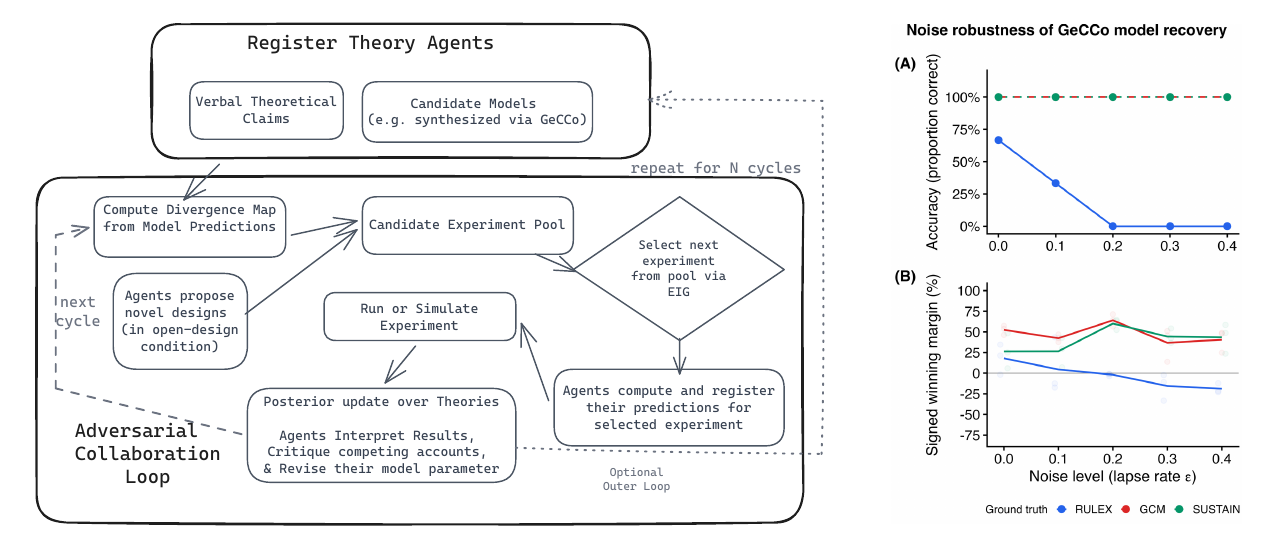}
    \caption{Left panel: Adversarial collaboration loop for theory adjudication and open experiment design. Right Panel: (A) Recovery accuracy across lapse rates for data generated by RULEX, GCM, and SUSTAIN. (B) Signed winning margins show that GCM and SUSTAIN remain more separable than RULEX.
    }
    \label{fig:debate_protocol}
    \vspace{-2mm}
\end{figure*}

Once models are registered, theory agents enter an adversarial debate cycle. Each cycle begins with a computational analysis of divergences in model predictions. Agents propose novel experiments based on the summarized divergence map, and proposals that satisfy structural validity checks are added to the pool. The system then uses the expected information gain (EIG, \cite{cavagnaro2010adaptive,myung2013tutorial}) to select the most informative experiment to run next. In this simulation study, the experimental outcomes are generated by a known cognitive model, but, in principle, they can easily be derived from cognitive foundation models or real human participants. Once predictions are registered and experimental outcomes are observed, the posterior beliefs about the theories are updated; the loop then returns to an LLM-mediated phase: the agents interpret the results, critique competing accounts, and may revise parameters and auxiliary claims.

After the final cycle, the system aggregates the experimental outcomes and debate traces to assess which theoretical claims and model instantiations are best supported by the accumulated evidence. In the current version, the resulting evidential record can be fed back into GeCCo to generate fresh candidate models for subsequent rounds.

\section{Results}
To evaluate this framework, we considered a synthetic setting, where we have full access to ground-truth theories, their model instances, and simulated data. Specifically, we considered three theory families from the human categorization literature: Generalized Context Model (GCM; \cite{nosofsky2011gcm}), Rule plus Exception (RULEX; \cite{nosofsky1994rule}), and Supervised and Unsupervised STratified Adaptive Incremental Network (SUSTAIN; \cite{love2004sustain}). 
GeCCo was instantiated with the Sonnet-4 model from Anthropic. In three separate experiments using each theory as the ground-truth, we generated synthetic item-level behavioral data with four levels of noise and tasked the system with recovering the underlying theory.  


Figure 1 summarizes the framework's recovery rate of the true generating theories under different noise levels, with noisy data created using a lapse-rate policy (with probability $\epsilon$, the response is random) for $\epsilon$ ranging from 0 to 0.4. We measured recovery performance by computing the proportion of runs in which the correct theory family was recovered and by the signed winning margin between the two main competing models. Overall, we find that our framework can perfectly recover models underlying the three ground-truth theories in the noiseless condition and to varying extents in the noisy condition, with accuracy depending on the theory type.

Specifically, we find that the GCM was perfectly recovered across the different noise settings, while SUSTAIN remained highly recoverable, except at the highest noise level (see Figure 1A). In contrast, RULEX's recoverability rapidly degraded and was no longer recoverable for $\epsilon>0.1$. A similar pattern emerges for the winning margins in Figure 1B: while GCM and SUSTAIN remained positively separated from competing models over much of the noise range, the margins for RULEX shrank rapidly and became negative.  These results indicate a potential bias in GeCCo for generating GCM and SUSTAIN-style models, but the system can still overcome these biases and find the ground-truth theory, as long as the underlying signal is strong. 

Taken together, these results provide the first proof of concept for an in-silico, closed-loop theory adjudication system for cognitive science, over open model and experimental design space. While the framework recovers multiple theory families in controlled settings, recovery was uneven across model classes, with persistent fragility for RULEX. This suggests that success depends on the inherent recoverability of synthesized candidate models as much as on the adjudication loop. Therefore, this is not yet a complete solution to automated theory building, but a step towards a system that tightly links model synthesis and experiment design. In future work, we plan to replace hand-crafted cognitive models with a more powerful simulator of human behavior \citep{binz2025foundation} and, eventually, real human behavior.
\printbibliography

\end{document}